\documentclass[runningheads]{llncs}

\usepackage{graphicx}
\usepackage{xstring}
\graphicspath{ {./images/} }
\usepackage{caption}
\usepackage{subcaption}
\usepackage{amsmath}
\usepackage{tikz}
\usepackage{comment}
\usepackage{hyperref}
\usepackage{booktabs}

\begin{document}
\title{Denoising ESG: quantifying data uncertainty from missing data with Machine Learning and prediction intervals}
\titlerunning{Denoising ESG}
%
\author{Sergio Caprioli \inst{1}  \and
 Jacopo Foschi \inst{2} \and 
 Riccardo Crupi \inst{2} \orcidID{0009-0005-6714-5161} \and
 Alessandro Sabatino \inst{2} \orcidID{0000-0002-1336-2057}
}
\authorrunning{}
%
\institute{Intesa Sanpaolo S.P.A., Milano MI 20121, Italy
\email{sergio.caprioli@intesasanpaolo.com}\\
\and
Intesa Sanpaolo S.P.A., Torino TO 10121, Italy\\
\email{\{name.surname\}@intesasanpaolo.com}}

\maketitle              
%


\begin{abstract}
Environmental, Social, and Governance (ESG) datasets are frequently plagued by significant data gaps, leading to inconsistencies in ESG ratings due to varying imputation methods. This study addresses the missing data issues in ESG datasets using machine learning techniques, comparing K-Nearest Neighbors,  Gradient Boosting, Multiple Imputation by Chained Equations (MICE) and Neural Networks. We focus on  quantifying the risk induced by data anomalies and provide tools to assess the impacts of this risk on the variability of the scores. By introducing prediction uncertainty using methods such as Predictive Mean Matching and Local Residual Draw, in order to assign confidence measures to individual predictions, we provide a nuanced understanding of prediction uncertainty. Empirical analyses show that these methods improve imputation accuracy and quantify uncertainty, which is required for reliable ESG scoring in banking and finance. 

\end{abstract}
\section{Introduction}
The growing attention towards the ramifications of climate change has bolstered international cooperation on sustainable finance. This collaboration involves initiatives from industry and institutions, as well as recommendations from regulators and legislators. There is a growing emphasis on integrating ESG criteria into the strategies and operations of banks, accompanied by the requisite development of adequate support tools.

European regulators acknowledge that ESG ratings play an important role in global capital markets, as investors, borrowers and issuers increasingly use those ESG ratings as part of the process of making informed, sustainable investment and financing decisions \cite{eu_agreement}. For this reason \begin{quote}
    ``Better comparability and increased reliability of ESG ratings would enhance the efficiency of that fast-growing market, thereby facilitating progress
towards the objectives of the Green Deal".
\end{quote} 
This aspect reverberated also on the banks, given that, as stated in \cite{EBA_consultation}, \begin{quote}
    
``Institutions should embed ESG risks in their regular processes including risk appetite, internal controls and ICAAP. Besides, institutions should monitor ESG risks through effective internal reporting frameworks and a range of backward and forward-looking ESG risks metrics and indicators."
\end{quote}

Given the importance attributed to the ESG metrics for the assessment of firms' performances on a given ESG issue, the topic of the reliability of such metrics plays a crucial role. Berg et.al \cite{berg2022aggregate} analyzed the divergence of ESG ratings based on data from six agencies, decomposing the divergence into contributions of scope, measurement and weight. They estimated that measurement contributes 56\% of the divergence. 
The issue assumes even a greater impact when considering that Banks should measure ESG risk for a variety of counterparties aggregating information derived from different sources: e.g. raw data from different rating agencies, internal sources, questionnaires. 

As part of the measurement divergence issue, datasets used to calculate ESG Scores are affected by vast data gaps, which are imputed by researchers and analysts with different methods, contributing to the inconsistency of ESG ratings (\cite{berg2022aggregate}, \cite{billio2024unpacking}). There is lack of regulation and standardization on ESG data disclosure by companies, which not only causes high missing rates, but also implies inconsistency on how different companies disclose data (e.g., companies provide ESG data for different Key Performance Indicators), with smaller companies typically being the less disclosing ones (\cite{kotsantonis2019four}, \cite{licari2021esg}). Also:  \begin{quote}``The number of companies covered by major ESG score providers typically ranges between 1,000 to 10,000, representing a major challenge for organizations with many more firms in their portfolios\cite{licari2021esg}.” \end{quote}

Then, imputation of missing ESG data plays a major role in determining companies’ ratings and in differentiating ESG Scoring results across different modelers. Data imputation approaches can be classified, in the context of ESG data, as\cite{kotsantonis2019four}:
\begin{itemize}
  \item	Rule-based methods: arbitrary imputation of missing datapoints based on an ad-hoc rule, either based on expert judgment or reflecting certain risk appetite  .
\item	Input-output models: imputation of the ESG metrics based on industrial and economic data (e.g., model developed by Licari et al. (2020)\cite{licari2021esg}).
\item Statistical approaches: data points are imputed by means of statistical and machine learning models.
\end{itemize}

While, on the one hand, the first approach can be tailored to meet specific judgments or goals of users of ESG ratings (e.g., assigning a penalty to companies not providing data or adopting a conservative approach), the second and third groups of approaches could help to achieve higher likelihood of imputed ESG data and final ESG ratings.

As reported in \cite{kotsantonis2019four}: \begin{quote}
    ``The differences in the imputation methods used by ESG researchers and analysts to
deal with vast ‘data gaps’ that span ranges of companies and time periods for different
ESG metrics can cause large ‘disagreements’ among the providers, with different gapfilling
approaches leading to big discrepancies." \end{quote}

The mentioned aspect also plays a critical role in the definition of internal ESG scoring, wherein banks are required to aggregate information from various sources. Hence, it is important to assess the reliability of the scores attributed to the counterparties taking into account the uncertainty induced by the anomalies in the data. 

The aim of this work is to quantify the uncertainty in ESG scores caused by missing data, leveraging probabilistic machine learning models that allow for estimating prediction intervals for imputed data instead of providing only a single point estimation. In this manner, the risk of incorrect classification in terms of the score attributed to the counterparty, due to the high incidence of missing data, is expressed by the dispersion of the prediction distribution estimated for the imputed data. This information allows for a clearer assessment of ESG scores, which is crucial for comparative analyses.2

\section{Handling missing data}

\subsection{Data} \label{data}
The case study has been implemented on Intesa Sanpaolo (ISP) ESG scoring dataset, as of March 2023. The dataset contains ESG data on 181071 companies belonging to ISP’s portfolio. The ESG dataset contains values of 155 KPIs measuring companies’ performances towards sustainability. KPIs are categoriezed into 22 Descriptors, which corresponds to 22 main topics determining the three ESG Pillars and the overall ESG score. The series of consecutive aggregation from KPIs to Descriptors, Pillars and ESG Score is achieved by means of weighted sums. Weights are defined by the ESG scoring model.

ISP’s ESG dataset is affected by a large and complex missing data problem, with high missing rate, with feature missingness depending on values (MAR mechanism) and/or missingness (cross-correlation of missingness indicators) of other features.
 
Gopal et al. (2020)\cite{tepelyan2023generative} suggests that most of missing ESG data follow Missing-at-Random (MAR) and Missing-not-at-Random (MNAR) mechanisms. Data are MNAR when the probability of being missing is dependent on the unobserved variables. In the MAR case, the probability of being missing is dependent on observed variables but independent of the unobserved variable. More rarely, the missing mechanism is Missing-Completely-at-Random (MCAR) when the probability of being missing is independent of both the observed and unobserved variables. ESG data are suspected of being MAR because relevance of some features (i.e., ESG KPIs) are not relevant given the industry sector and MNAR when companies might withhold their data when worse than industry benchmark. 

\subsection{Analysis of a set of machine learning imputation methods} \label{imputation_methods}

The current study compared several statistical and machine learning methods with well-established application in the data imputation field, being: K-Nearest Neighbors \cite{pujianto2019k,zhang2012nearest}, Multiple Imputation Chained Equations (MICE) \cite{van2018flexible}, Histogram Gradient Boosting (HGB) \cite{guryanov2019histogram}, Denoising Autoencoders \cite{pereira2020reviewing} and Graph Convolution Networks (GCN) \cite{spinelli2020missing}. 

KNN and HGB were used as multi-input single-output regressors, imputing each KPI using all the others as predictors. KNN estimates the target variable by averaging the values of its K-nearest neighbors in the feature space. The neighbors are defined based on their proximity to the query point in the feature space. In this study, the Euclidean distance metrics was used. The pairwise Euclidean distance between samples was calculated only on KPIs which are not missing in both samples, thus allowing to naturally handle missing data. The main advantage of the KNN is its non parametric nature, which allows not to make assumptions on the regression model while preserving flexibility in handling complex relationships. HGB is an ensemble learning technique that combines the boosting algorithm \cite{ferreira2012boosting} with histogram-based decision trees, where each tree is built using histograms to efficiently split data into intervals. It offers increased computational efficiency and improves predictive performance compared to traditional gradient boosting. In this study, we used the Scikit-Learn implementation \cite{pedregosa2011scikit}, which has native support for missing values (as further detailed in Paragraph \ref{mi_workflow}).

The MICE algorithm is an iterative method for handling missing data, where incomplete values are imputed by sequentially regressing each missing variable on the other observed variables, iterating until convergence. In this study, a Random Forest (RF)\cite{breiman2001random} regression model was used in each iteration and the algorithm was used in a 'single imputation' mode. After an initial random initialization of missing values by drawings from empirical marginals of features, point estimates from the RF was used at each iteration to impute missing data points, until convergence to stable values of data points. As later described  paragraph 2.3, MICE can be used in an enhanced algorithm to generate multiple imputations and account for uncertainty.

DAE are neural network models trained to reconstruct clean data from corrupted inputs by learning a compact representation of the data through an encoder network, followed by a decoder network that reconstructs the original input, enabling effective noise removal. In case of missing data imputation, missing data coincides with the input noise to be removed. Differently from above mentioned methods, DAE is a multi-input multi-output model, which reconstruct the clean set of KPIs starting from the corrupted set as input.

GCNs are neural network architectures designed to process and analyze data structured as graphs by performing convolutions directly on the graph structure, enabling information propagation and feature extraction across nodes. This model firstly requires the transformation of our tabular dataset into its graph representation. This was achieved by representing samples as nodes of the graph, KPIs as nodes features and edge weights (i.e., 'length' of edges linking nodes) as Manhattan pairwise distances between samples. Manhattan distances were calculated by weighting the contribution of each KPI by its weight in determining the total ESG score (weight value is a-priori determined by the ESG scoring model). Edges with weights lower than the 95\textsuperscript{th} percentile were removed, in order to achieve a sparse graph were only closer relationships among nodes (i.e., samples) are represented. Once the graph was constructed, a GCN made of 2 encoding and 2 decoding layers was trained on it. This methodology follows the approach described in \cite{spinelli2020missing}.  

\begin{figure}[h]
\centering
\includegraphics[height=0.7\linewidth]{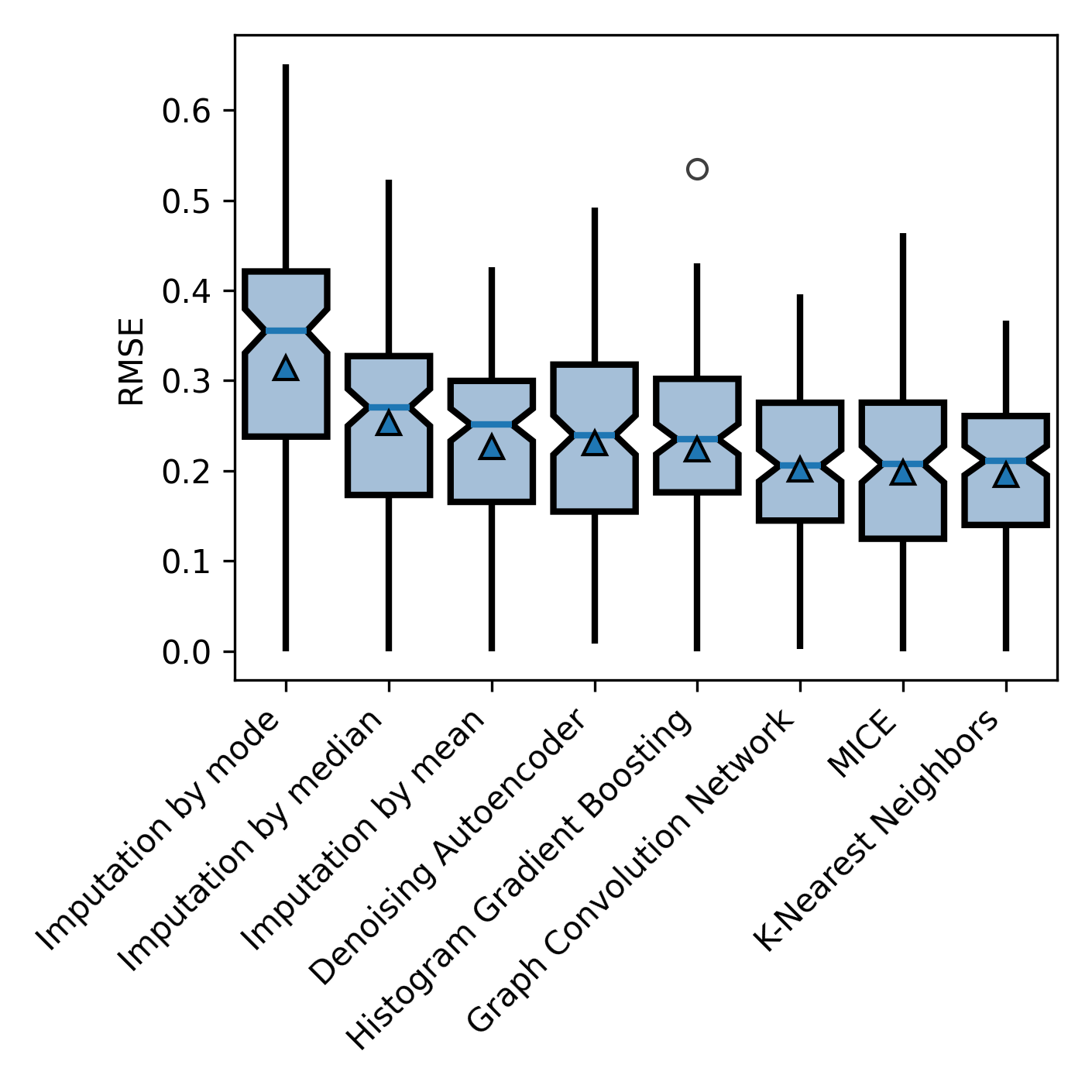}
\caption{Test Root Mean Squared Error (RMSE) of imputation methods. Boxplots account for RMSE variability among KPIs. Mean values are reported as red triangles.}
\label{fig:single_imputation}
\end{figure}

All the above mentioned methods were benchmarked against simple imputation by mean, mode and median. The comparison is performed on a subset of 505 counterparty belonging to Tier 1 of the ESG dataset. 30\% of data were sampled out for testing. An additional 30\% of available data points were removed to the purpose of test performance measurement.  
As reported in Figure \ref{fig:single_imputation} KNN, MICE and GCN are the only three methods which provide any, even if small, improvement in imputation performance with respect to the simple imputation by means. As expected, all other simple imputation approaches are outperformed. Results are consistent with the literature, where KNN and MICE are two of the most used data imputation methods and often perform similarly to more complex machine learning based methods \cite{pereira2020reviewing} \cite{you2020handling}. Also, similar performances between the KNN and the GCN are consistent with the GCN characteristics. The GCN model is trained on a graph representation of the dataset which connects samples with their pairwise Manhattan distance. The GCN can be then be interpreted as a generalization of the KNN, with more flexible identification of observation neighborhoods and aggregation rules, thanks to the Convolution Network. GCN, KNN and MICE better performances can be also due to their native ability to handle fully corrupted datasets, i.e. dataset where no samples are available without any missing data point, like in current case study.  In the case of GCN and KNN, pairwise distances between samples are calculated considering only common available features (i.e., KPIs), while the MICE algorithm mitigate the effect of the initial imputation of missing data by means of the multiple iterations of imputations, which converge to stable values. Differently, DAE \cite{pereira2020reviewing} need an initial pre-imputation of missing values, which could bias training and penalize its performances. This could be among reasons why DAE performs even worse than a simple imputation by mean.  

\subsection{Multiple Imputation workflow} \label{mi_workflow}
The limited improvement of imputation accuracy of models tested in paragraph \ref{imputation_methods} suggests that while regression and machine learning models can be a good option for ESG data imputation, still the uncertainty introduced by the presence of missing data and by the imputation model itself should be estimated, too. Also, both imputation by statistics of observed data, like the mean, and by regression or machine learning models, like in the approaches reviewed in paragraph \ref{imputation_methods}, may lead to biased estimates of quantities based on imputed data and underestimation of data variability. Imputation by statistics, like the mean, usually implies underestimation of variance, distortion of relations between variables and bias of most of estimates based on imputed data. Also, regression and machine learning imputation usually lead to upwards biased correlations and systematically underestimated variability of imputed data. These drawbacks are solved by the Multiple Imputation (MI) approach, which creates multiple complete datasets, by replacing the missing values by plausible data values. These plausible values are drawn from a distribution specifically modeled for each missing entry:
\begin{quote}
"The [multiple] imputed datasets are identical for the observed data entries, but differ in the imputed values. The magnitude of these difference reflects our uncertainty about what value to impute\cite{van2018flexible}."
\end{quote}

Importantly, MI separates the solution of the missing data problem from the solution of the complete data problem. First, missing data are estimated and, second, the analysis of interest (in this case, the calculation of ESG Scores) is performed on each complete dataset. Complete datasets and results of the analysis can be pooled to estimate their probability distribution, which incorporate uncertainty introduced by the presence of missing data and the imputation model itself.

The MICE was selected to model the conditional distributions of missing data, since it is one of the best-established methods in the imputation field and it resulted in one of best predictive performances in current case study.
RF are selected as regression model to integrate in the MICE algorithm and prediction uncertainty was introduced in RF estimates by means of Predictive Mean Matching (PMM)\cite{doove2014recursive} and Local residual Draw (LRD)\cite{morris2014tuning}. These two methods, which both come from the family of 'Hot Deck' methods, draw random predictions from a set of n 'donors', identified as the n closest observations to the predicted variable. PMM keeps the drawn observation (the 'donor') as the predicted value. Instead, LRD sums the residuals associated with each donor to the predicted value. The use of RF as predictive model enabled us to use out-of-bag\cite{breiman1996out}  residuals to obtain a better estimation of the predictive interval. Both PMM and LRD aims at accounting for prediction uncertainty and allows to simulate prediction intervals. Also, Multiple Imputation by MICE can be seen as a Markov Chain Monte Carlo process, specifically as a Gibbs sampler. As pointed out in Morris et al. (2014)\cite{morris2014tuning}, use of PMM and LRD is typically motivated by the notion that they provide a degree of robustness when the imputation model is misspecified, for example if the normality assumption is in question, residuals are heteroscedastic, or associations are nonlinear.".

An ad-hoc methodology was developed to both perform the Multiple Imputation and calculate a robust test measure of its performance. Data imputation models in the literature are often evaluated by taking out complete samples from the original dataset and devoting a fraction of these data to test. A fraction of data points is removed randomly to simulate missingness and the imputation model is tested against true values. This framework is easy to apply but does not fit the characteristics of this imputation problem. As mentioned in paragraph \ref{data}, ESG data are often MAR (and not MCAR), thus testing the imputation model on randomly removed data would not reproduce actual working condition of the model. Based on the discussion above, the following 5-steps workflow was developed: 

1.	MICE imputation (first run): multiple complete datasets are generated starting from the raw dataset with missing data. Diagnostics on imputation is performed by checking  distributional discrepancy between observed and imputed data.

2.	Calibration of a missing data simulation model: calibration of one model for each KPI to predict its missingness probability given values of all other KPIs. HGB (see paragraph \ref{imput_methods} wasselected to this purpose. Importantly, the implementation of HGB natively handle missing information on predictors, thus leveraging its explanatory power along with available values of KPIs. Specifically, during training the tree grower learns at each split point whether samples with missing values should go to the left or right child, based on the potential gain on performance. When predicting, samples with missing values are assigned to the left or right child consequently. When the missingness pattern is predictive, the splits can be performed on whether the feature value is missing or not. This feature of the HGB implementation is fundamental since it allows the model to capture the complete missingness mechanism described in the exploratory analysis, where occurrence of missing data for KPIs appeared to be correlated to both other KPIs values (MAR mechanism) and missingness indicators  of other KPIs.

3.	Augmentation: the sequence of RF trained in step 1 of the workflow is used to generate a synthetic version of one of datasets imputed in step 1. In doing this, RF models are used to replace values of each KPI with their predictions. It must be reiterated that, as for step 1, predictions are drawn from KPI distribution conditional on other KPIs values, by coupling RF point prediction with PMM and LRD (see description at the first section of this paragraph). Therefore, this step result in a new synthetic dataset where all data points will differ from the original one, while maintaining the same multivariate distribution of KPIs (i.e., preserving marginal distributions and cross-correlations between KPIs).

4.	Amputation: models trained in step 2 are used to simulate the probability of missingness of each data point of the synthetic dataset from step 3. For each data point of a KPI, probability of missingness is predicted based on other KPIs values. The actual missingness of the data point is then determined by drawing from a Bernoulli variable with probability of occurrence equal to the probability of missingness. The simulation is repeated for each KPI and the full round of simulations on all KPIs is iterated 10 times, too, to reach convergence to stable distributional characteristics of missing data. This step result in a synthetic dataset which maintains distributional behavior of both the observed values and missing data occurrence of the real ESG dataset which is the input of step 1.

5.	MICE (second run): step 1 is repeated on a training subset (70\%) of synthetic data from step 4. Trained MICE is then used to input the test subset (30\%) and performance metrics are calculated. Mean Absolute Error (MAE), Root Mean Squared Error (RMSE), Coverage Rate (CR) and Average Width (AW) were calculated, both at KPI, Descriptor, Pillar and overall ESG level. Since the main goal of multiple imputation is the reconstruction of the distribution of data, we focused on CR as most important metrics. CR is defined as the fraction of data points falling withing their 95\% prediction interval, estimated from multiple imputations. We expect a good imputation model to have a CR which is as close as possible to the nominal prediction interval rate (95\%). Getting good CR is then a fundamental premise to achieve a good approximation of ESG score distribution. AW is defined as the average range of the prediction interval.

Within step 1, the MICE algorithm was run 50 times to generate 50 complete datasets. Descriptor, Pillar and ESG score were calculated on each complete dataset, resulting in n estimates of each indicator, and, therefore, in an estimate of their distribution for each counterparty. The sequence of augmentation, “amputation” and imputation of steps from 2 to 5 resulted in the estimate of the performance of the MICE algorithm: the MI by MICE was repeated on a synthetic dataset, which shares same statistical characteristics of the real one and for which we know values of missing data points. The whole workflow was repeated for both PMM and LRD. 

\subsection{Assessment of ESG score uncertainty due to missing data}

\begin{figure}
    \centering
    \includegraphics[width=1\linewidth]{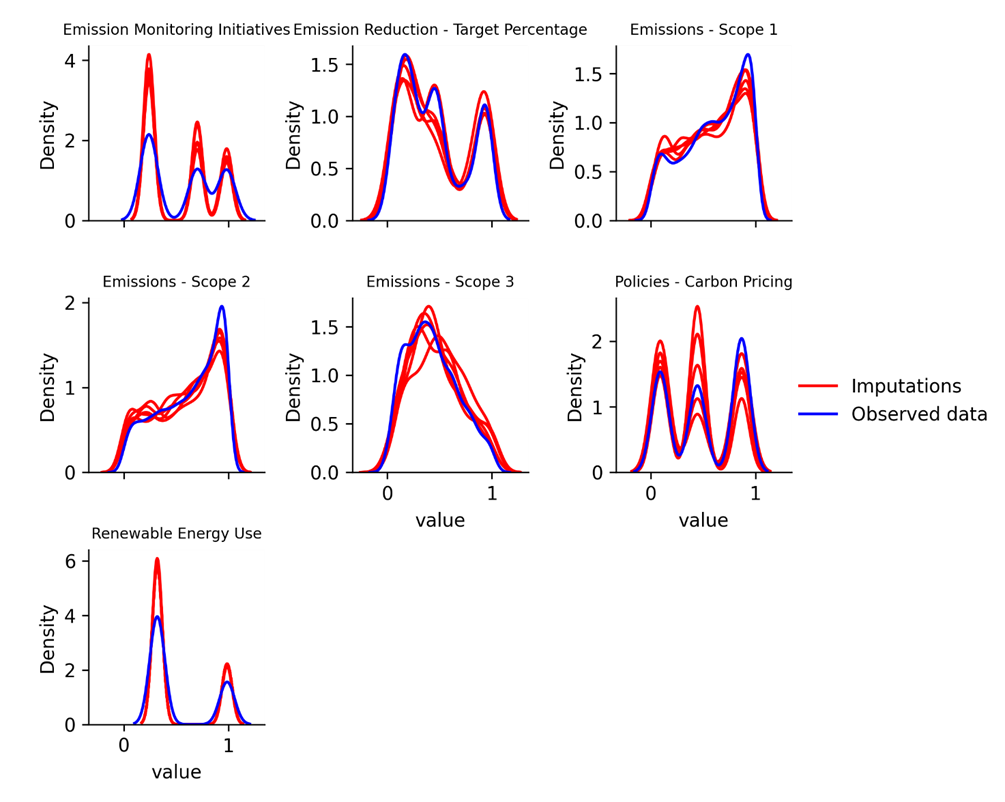}
    \caption{Comparison of marginal distributions of observed values and imputed values from 5 imputations of KPIs belonging to KPI "Carbon Footprint".}
    \label{fig:figure_diagnostics_mcar}
\end{figure}

Diagnostics of MI results revealed good matching between distributions of observed and imputed values. An example is reported in Figure \ref{fig:figure_diagnostics_mcar}, showing such comparison for KPIs belonging to Descriptor "Carbon Footprint", for imputations by PMM method. As can be seen, no dramatical distributional discrepancies can be detected. Also, the use of PMM effectively handle the several semi-continuous variables within the ESG dataset (see KPIs "Emission Monitoring Initiatives", "Policies - Carbon Pricing" and Renewable Energy Use"), by mapping the imputed values on the actual support of the variable. 

Table \ref{tab:mi_performance} reports performance metrics on Multiple Imputation both when using PMM and LRD, where Coverage Rate is the key indicator. CRs equal to 89\% and 93\% on pillar and ESG score estimates were calculated for PMM and LRD, respectively, suggesting that the Multiple Impitation by MICE algorithm is a robust approach, with negligible probability of underestimation of companies ratings. 

\begin{table}[h!]
\centering
\caption{Root Mean Squared Error (RMSE), Mean Absolute Error (MAE), Coverage Rate (CR) and Average Width (AW) calculated on multiple imputation by MICE using both Predictive Mean Matching (PMM) and Local Residual Draw (LRD). Results are reported both at the ESG and Pillar Score level.}
\begin{tabular}{c | cc | cc | cc | cc}
\toprule
     &  \multicolumn{2}{c}{ESG} &  \multicolumn{2}{c}{Environment} &  \multicolumn{2}{c}{Social} &  \multicolumn{2}{c}{Governance}\\
\midrule
    Metrics &   PMM &  LRD &   PMM &  LRD &   PMM &  LRD &   PMM &  LRD\\
\midrule
       RMSE &  0.04 & 0.04 &  0.06 & 0.06 &  0.04 & 0.05 &  0.07 & 0.07\\
        MAE &  0.03 & 0.03 &  0.05 & 0.05 &  0.03 & 0.03 &  0.06 & 0.05\\
    CR (\%) &  89.7 & 92.2 &  91.1 & 91.3 &  89.0 & 93.7 &  92.0 & 89.9\\
         AW &  0.12 & 0.14 &  0.18 & 0.21 &  0.12 & 0.16 &  0.23 & 0.22\\
\bottomrule
\end{tabular}
\label{tab:mi_performance}
\end{table}

\begin{figure}
    \centering
    \includegraphics[width=1\linewidth]{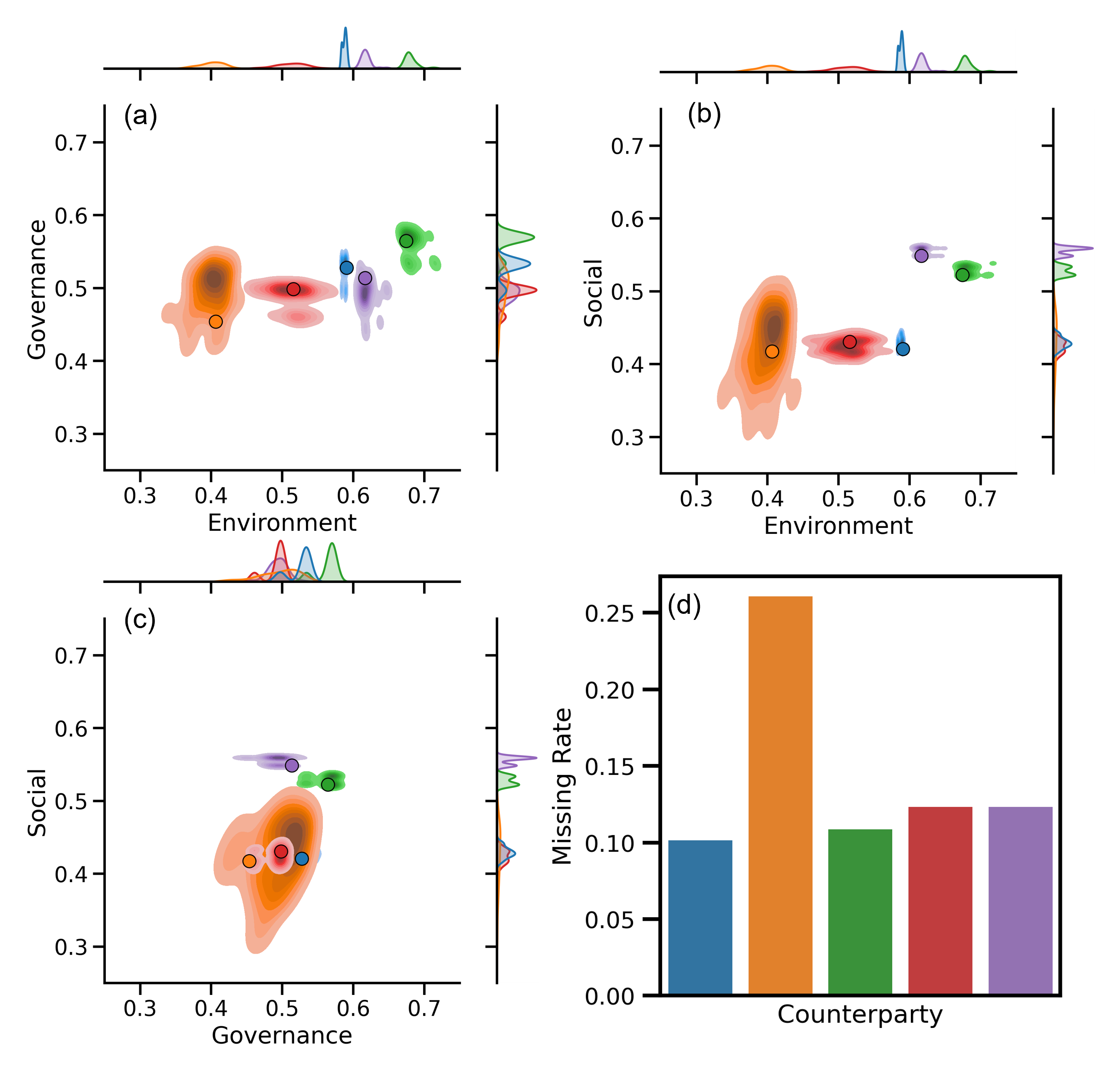}
    \caption{Jointplot comparing distributions Pillar Scores from multiple imputations by MICE with Pillar Scores from single imputation by MICE (a, b and c). Results from 5 example counterparties are reported in different colors. Missing rate of each counterparty is reported in subplot d.}
    \label{fig:mi_results}
\end{figure}

\begin{figure}
    \centering
    \includegraphics[width=.7\linewidth]{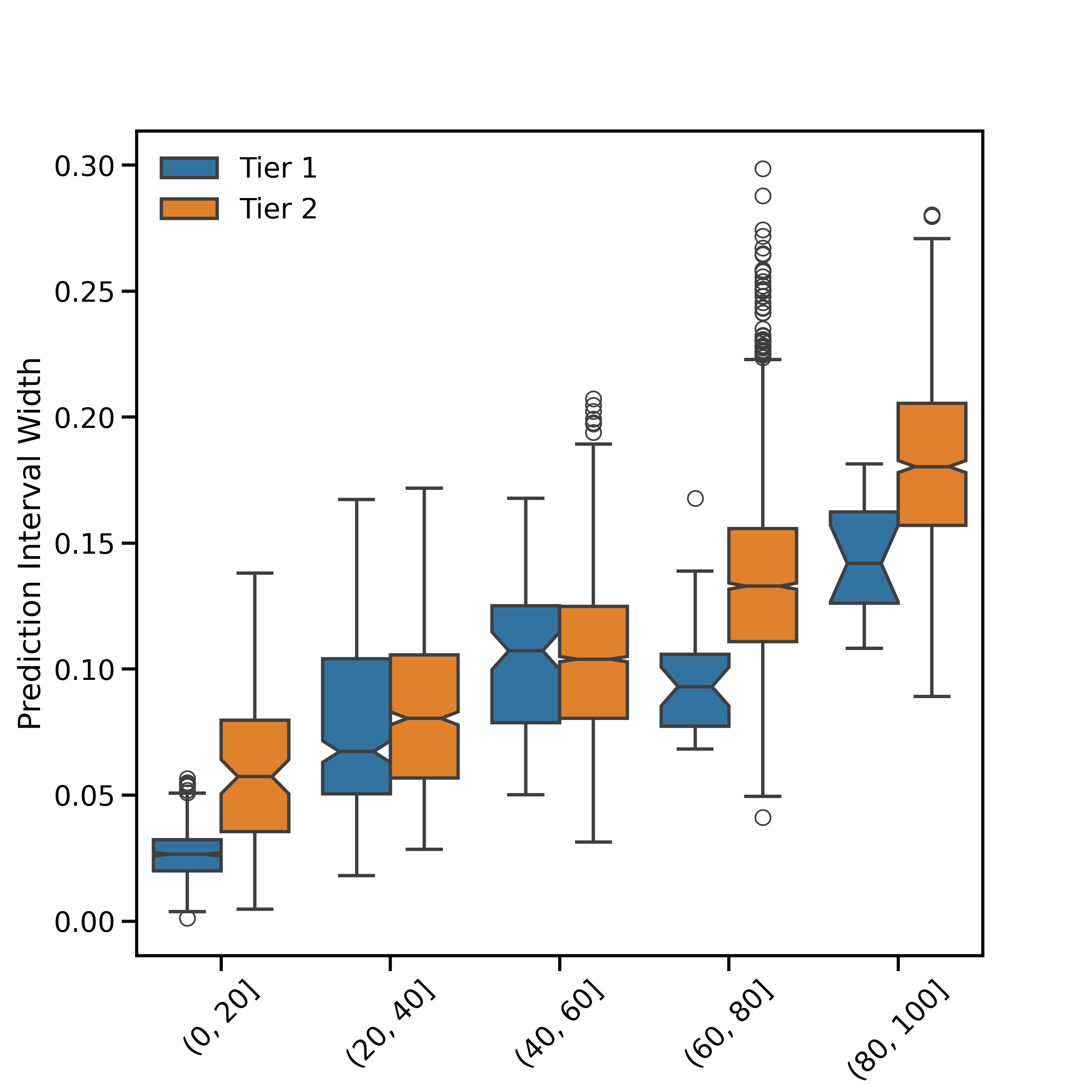}
    \caption{Boxplot of width of prediction intervals of ESG Scores of all counterparties, by Tier and missing rate (i.e., proportion of missing KPIs pee counterparty)}
    \label{fig:pi_width}
\end{figure}

Figure \ref{fig:mi_results} gives an example of estimated distributions of Pillar Scores for 5 counterparties within the case study dataset, comparing them with the point estimate of scores we could obtain by using the MICE algorithm in a "single" imputation approach (for disclosure reasons, the data in the figure are represented by colors instead of company identifiers). It is evident how this methodology propagates uncertainty proportionally to mising rates: results on the counterparty represented in orange corresponds to orignial data with 27\% missing rate, the highest among the 5 shown and the resulting distribution covers the largest Pillar Score space. Comparison of this distribution with the point estimate we would get from a single imputation stresses the dramatical difference between the two approaches. Importantly, in this example we see that, due to too high missing rate, we can not distinguish the Governance score of counterparty in orange from counterparties in red and blue. Counterparties in red and purple have a 12\% missing rate, with a consequently smaller variability. Importantly, since the imputation model act at the KPI level, the MI approach allows to uderstand which Pillar (and, furtherly breaking down, Descriptors) is affected by higher uncertainty. While results on the counterparty in red are equally uncertain for all pillars, uncertainty lays mostly in the Governance Pillar score for the counterparty in purple. Finally, counterparties in blue and green have a 10\% missing rate, with the lowest resulting uncertainty among the five example counterparties. The relationship between the width of ESG Score prediction interval and missing rate is evident in Figure \ref{fig:pi_width}. Also, this shows that, for the same missing rate bin, prediction intervals tend to be wider for Tier 2 with respect to Tier 1, suggesting that not only missing rate, but also data quality helps the imputation model to be more precise (Tiers are identified by data availability, with higher Tiers being covered by "top tier" global vendors). 

\clearpage++

\section{Conclusions}
The presented study provides an insight on possible tools to handle the missing data problem in ESG datasets. Use of well established techniques like KNN and MICE can improve imputation accuracy with respect to simple imputation by mean. Such algorithms are widely implemented in several packages (\cite{van2011mice}, \cite{pedregosa2011scikit}). GCN lead to similar accuracy, but they require higher computational effort, both to build the graph representation of data and to train the neural network model. Further research is then required to possibly better fine-tune the GCN model and achieve higher performances and justify the modeling and computational effort. Also, the study points at the Multiple Imputation by MICE as a good practice to keep into account uncertainty introduced by data imputation. Besides providing robust prediction intervals of missing data, it enables the propagation of missing data uncertainty in ESG scoring, thus integrating in a risk-perspective which is necessary in banking and finance applications. 

\section*{Disclaimer\label{sec:disclaimer}}
The views and opinions expressed within this paper are those of the authors and do not necessarily reflect the official policy or position of Intesa Sanpaolo. Assumptions made in the analysis, assessments, methodologies, models and results are not reflective of the position of any entity other than the authors.

\nocite{*}
\bibliography{bib/ref.bib}

\begin{thebibliography}{10}

\bibitem{eu_crd}
Directive 2013/36/eu of the {E}uropean {P}arliament and of the {C}ouncil.
\newblock {\em Official Journal of the European Union\/} (26 June 2013).

\bibitem{eu_regulation}
Regulation (eu) no 575/2013 of the {E}uropean {P}arliament and of the {C}ouncil.
\newblock {\em Official Journal of the European Union\/} (26 June 2013).

\bibitem{alabadla2022systematic}
{\sc Alabadla, M., Sidi, F., Ishak, I., Ibrahim, H., Affendey, L.~S., Ani, Z.~C., Jabar, M.~A., Bukar, U.~A., Devaraj, N.~K., Muda, A.~S., et~al.}
\newblock Systematic review of using machine learning in imputing missing values.
\newblock {\em IEEE Access 10\/} (2022), 44483--44502.

\bibitem{EBA_consultation}
{\sc Authority, E.~B.}
\newblock Draft guidelines on the management of esg risks.
\newblock {\em EBA/CP/2024/02\/} (2024).

\bibitem{berg2022aggregate}
{\sc Berg, F., Koelbel, J.~F., and Rigobon, R.}
\newblock Aggregate confusion: The divergence of esg ratings.
\newblock {\em Review of Finance 26}, 6 (2022), 1315--1344.

\bibitem{billio2024unpacking}
{\sc Billio, M., Fitzpatrick, A.~C., Latino, C., and Pelizzon, L.}
\newblock Unpacking the esg ratings: Does one size fit all?

\bibitem{bonacorsi2022esg}
{\sc Bonacorsi, L., Cerasi, V., Paola, G., and Manera, M.}
\newblock Esg factors and firms’ credit risk.

\bibitem{bondarenko2016graphical}
{\sc Bondarenko, I., and Raghunathan, T.}
\newblock Graphical and numerical diagnostic tools to assess suitability of multiple imputations and imputation models.
\newblock {\em Statistics in medicine 35}, 17 (2016), 3007--3020.

\bibitem{breiman1996out}
{\sc Breiman, L.}
\newblock Out-of-bag estimation.

\bibitem{breiman2001random}
{\sc Breiman, L.}
\newblock Random forests.
\newblock {\em Machine learning 45\/} (2001), 5--32.

\bibitem{cover1967nearest}
{\sc Cover, T., and Hart, P.}
\newblock Nearest neighbor pattern classification.
\newblock {\em IEEE transactions on information theory 13}, 1 (1967), 21--27.

\bibitem{doove2014recursive}
{\sc Doove, L.~L., Van~Buuren, S., and Dusseldorp, E.}
\newblock Recursive partitioning for missing data imputation in the presence of interaction effects.
\newblock {\em Computational statistics \& data analysis 72\/} (2014), 92--104.

\bibitem{ferreira2012boosting}
{\sc Ferreira, A.~J., and Figueiredo, M.~A.}
\newblock Boosting algorithms: A review of methods, theory, and applications.
\newblock {\em Ensemble machine learning: Methods and applications\/} (2012), 35--85.

\bibitem{guryanov2019histogram}
{\sc Guryanov, A.}
\newblock Histogram-based algorithm for building gradient boosting ensembles of piecewise linear decision trees.
\newblock In {\em Analysis of Images, Social Networks and Texts: 8th International Conference, AIST 2019, Kazan, Russia, July 17--19, 2019, Revised Selected Papers 8\/} (2019), Springer, pp.~39--50.

\bibitem{joenssen2012hot}
{\sc Joenssen, D.~W., and Bankhofer, U.}
\newblock Hot deck methods for imputing missing data: the effects of limiting donor usage.
\newblock In {\em Machine Learning and Data Mining in Pattern Recognition: 8th International Conference, MLDM 2012, Berlin, Germany, July 13-20, 2012. Proceedings 8\/} (2012), Springer, pp.~63--75.

\bibitem{RF_conformal}
{\sc Johansson, U., Bostr{\"o}m, H., L{\"o}fstr{\"o}m, T., and Linusson, H.}
\newblock Regression conformal prediction with random forests.
\newblock {\em Machine learning 97\/} (2014), 155--176.

\bibitem{kotsantonis2019four}
{\sc Kotsantonis, S., and Serafeim, G.}
\newblock Four things no one will tell you about esg data.
\newblock {\em Journal of Applied Corporate Finance 31}, 2 (2019), 50--58.

\bibitem{landerman1997empirical}
{\sc Landerman, L.~R., Land, K.~C., and Pieper, C.~F.}
\newblock An empirical evaluation of the predictive mean matching method for imputing missing values.
\newblock {\em Sociological methods \& research 26}, 1 (1997), 3--33.

\bibitem{licari2021esg}
{\sc Licari, J., Loiseau-Aslanidi, O., Piscaglia, S., and Gonzalez, B.~S.}
\newblock Esg score predictor: Applying a quantitative approach for expanding company coverage.
\newblock {\em Moody’s Analytics\/} (2021).

\bibitem{morris2014tuning}
{\sc Morris, T.~P., White, I.~R., and Royston, P.}
\newblock Tuning multiple imputation by predictive mean matching and local residual draws.
\newblock {\em BMC medical research methodology 14\/} (2014), 1--13.

\bibitem{eu_agreement}
{\sc of~the~{E}uropean {U}nion, C.}
\newblock Proposal for a regulation of the {E}uropean {P}arliament and of the {C}ouncil on the transparency and integrity of {E}nvironmental, {S}ocial and {G}overnance ({ESG}) rating activities, and amending regulation ({EU}) 2019/2088.

\bibitem{pedregosa2011scikit}
{\sc Pedregosa, F., Varoquaux, G., Gramfort, A., Michel, V., Thirion, B., Grisel, O., Blondel, M., Prettenhofer, P., Weiss, R., Dubourg, V., et~al.}
\newblock Scikit-learn: Machine learning in python.
\newblock {\em the Journal of machine Learning research 12\/} (2011), 2825--2830.

\bibitem{pereira2020reviewing}
{\sc Pereira, R.~C., Santos, M.~S., Rodrigues, P.~P., and Abreu, P.~H.}
\newblock Reviewing autoencoders for missing data imputation: Technical trends, applications and outcomes.
\newblock {\em Journal of Artificial Intelligence Research 69\/} (2020), 1255--1285.

\bibitem{pujianto2019k}
{\sc Pujianto, U., Wibawa, A.~P., Akbar, M.~I., et~al.}
\newblock K-nearest neighbor (k-nn) based missing data imputation.
\newblock In {\em 2019 5th International Conference on Science in Information Technology (ICSITech)\/} (2019), IEEE, pp.~83--88.

\bibitem{sahin2022environmental}
{\sc Sahin, {\"O}., Bax, K., Czado, C., and Paterlini, S.}
\newblock Environmental, social, governance scores and the missing pillar—why does missing information matter?
\newblock {\em Corporate Social Responsibility and Environmental Management 29}, 5 (2022), 1782--1798.

\bibitem{shah2014comparison}
{\sc Shah, A.~D., Bartlett, J.~W., Carpenter, J., Nicholas, O., and Hemingway, H.}
\newblock Comparison of random forest and parametric imputation models for imputing missing data using mice: a caliber study.
\newblock {\em American journal of epidemiology 179}, 6 (2014), 764--774.

\bibitem{spinelli2020missing}
{\sc Spinelli, I., Scardapane, S., and Uncini, A.}
\newblock Missing data imputation with adversarially-trained graph convolutional networks.
\newblock {\em Neural Networks 129\/} (2020), 249--260.

\bibitem{tepelyan2023generative}
{\sc Tepelyan, R., and Gopal, A.}
\newblock Generative machine learning for multivariate equity returns.
\newblock In {\em Proceedings of the Fourth ACM International Conference on AI in Finance\/} (2023), pp.~159--166.

\bibitem{van2018flexible}
{\sc Van~Buuren, S.}
\newblock {\em Flexible imputation of missing data}.
\newblock CRC press, 2018.

\bibitem{van2011mice}
{\sc Van~Buuren, S., and Groothuis-Oudshoorn, K.}
\newblock mice: Multivariate imputation by chained equations in r.
\newblock {\em Journal of statistical software 45\/} (2011), 1--67.

\bibitem{vovk2005algorithmic}
{\sc Vovk, V., Gammerman, A., and Shafer, G.}
\newblock {\em Algorithmic learning in a random world}, vol.~29.
\newblock Springer, 2005.

\bibitem{you2020handling}
{\sc You, J., Ma, X., Ding, Y., Kochenderfer, M.~J., and Leskovec, J.}
\newblock Handling missing data with graph representation learning.
\newblock {\em Advances in Neural Information Processing Systems 33\/} (2020), 19075--19087.

\bibitem{zaffran2023conformal}
{\sc Zaffran, M., Dieuleveut, A., Josse, J., and Romano, Y.}
\newblock Conformal prediction with missing values.
\newblock In {\em International Conference on Machine Learning\/} (2023), PMLR, pp.~40578--40604.

\bibitem{zhang2012nearest}
{\sc Zhang, S.}
\newblock Nearest neighbor selection for iteratively knn imputation.
\newblock {\em Journal of Systems and Software 85}, 11 (2012), 2541--2552.

\end{thebibliography}
\bibliographystyle{acm}

\newpage

\end{document}